\documentclass[runningheads]{llncs}
\usepackage{graphicx}

\usepackage{tikz}
\usepackage{comment}
\usepackage{amsmath,amssymb} 
\usepackage{color}

\usepackage{graphicx}
\usepackage{booktabs}
\usepackage{xcolor}
\usepackage{helvet}
\usepackage{courier}
\usepackage{balance}
\usepackage{textcomp,booktabs}
\usepackage{footmisc}
\usepackage[normalem]{ulem}
\usepackage{multirow}
\usepackage{setspace}
\usepackage{bigstrut}
\usepackage{array}
\usepackage{amsfonts}
\usepackage{bm}
\usepackage{url}
\usepackage{mathrsfs}
\usepackage{booktabs}
\usepackage{adjustbox}
\usepackage{caption}
\usepackage{cuted}
\usepackage{capt-of}
\usepackage{colortbl}
\usepackage[ruled,vlined]{algorithm2e}

\usepackage[accsupp]{axessibility}  

\usepackage{bbding}
\makeatletter
\def\@fnsymbol#1{\ensuremath{\ifcase#1\or \textsuperscript{~\Envelope}\or \ddagger\or
   \mathsection\or \mathparagraph\or \|\or **\or \dagger\dagger
   \or \ddagger\ddagger \else\@ctrerr\fi}}
\makeatother

\newcommand{\Rb}{\mathbb{R}}
\newcommand{\Dm}{\mathcal{D}}
\newcommand{\Lm}{\mathcal{L}}
\definecolor{Gray}{gray}{0.9}

\definecolor{Gray}{gray}{0.9}

\begin{document}
\pagestyle{headings}
\mainmatter

\title{SPCL: A New Framework for Domain Adaptive Semantic Segmentation via Semantic Prototype-based Contrastive Learning} 

\titlerunning{~}

\author{Binhui Xie \quad Mingjia Li \quad Shuang Li\thanks{Corresponding author.}}

\authorrunning{~}

\institute{School of Computer Science and Technology, Beijing Institute of Technology \\
\email{\{binhuixie, mingjiali, shuangli\}@bit.edu.cn}}

\maketitle

\begin{abstract}
Although there is significant progress in supervised semantic segmentation, it remains challenging to deploy the segmentation models to unseen domains due to domain biases. Domain adaptation can help in this regard by transferring knowledge from a labeled source domain to an unlabeled target domain. Previous methods typically attempt to perform the adaptation on global features, however, the local semantic affiliations accounting for each pixel in the feature space are often ignored, resulting in less discriminability. To solve this issue, we propose a novel semantic prototype-based contrastive learning framework for fine-grained class alignment. Specifically, the semantic prototypes provide supervisory signals for per-pixel discriminative representation learning and each pixel of source and target domains in the feature space is required to reflect the content of the corresponding semantic prototype. In this way, our framework is able to explicitly make intra-class pixel representations closer and inter-class pixel representations further apart to improve the robustness of the segmentation model as well as alleviate the domain shift problem. Our method is easy to implement and attains superior results compared to state-of-the-art approaches, as is demonstrated with a number of experiments. The code is publicly available at \url{https://github.com/BinhuiXie/SPCL}.
\end{abstract}

\section{Introduction}

\begin{figure*}
    \centering
    \includegraphics[width=\textwidth]{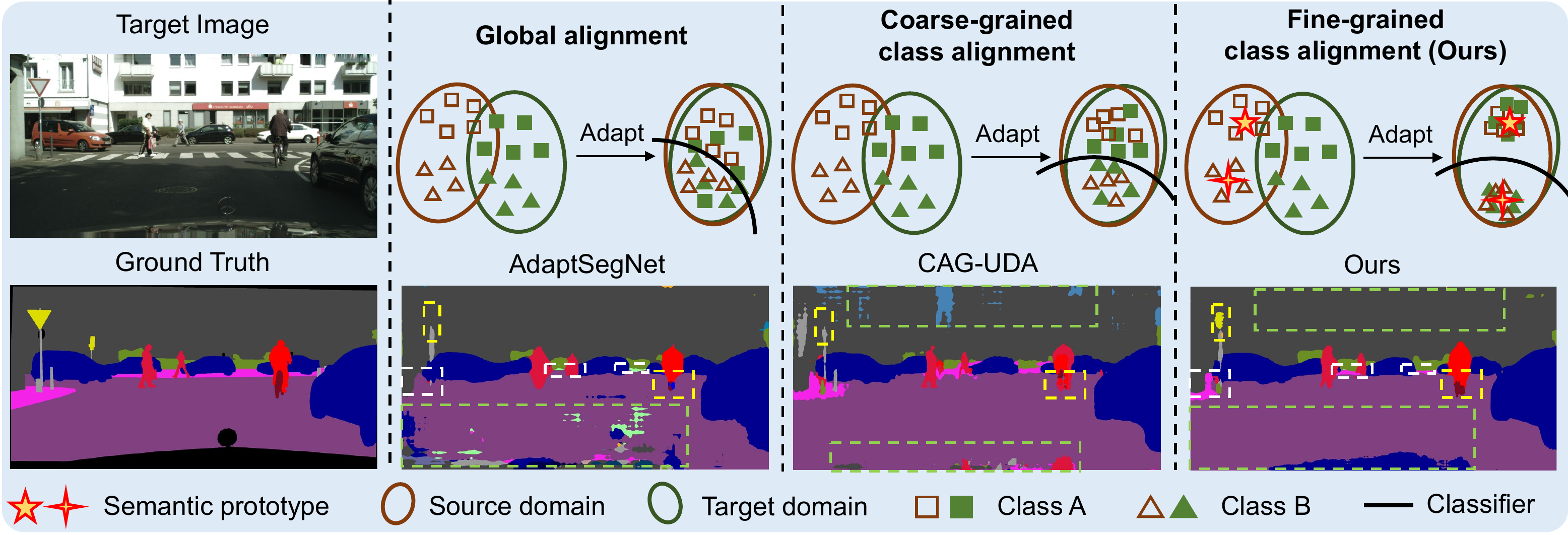}
    \caption{{\bf Main idea.} The mechanism and segmentation result of global alignment are shown in the second column, where many pixels are adapted into wrong area due to lack of local consistency. In the third column, the coarse-grained class alignment alleviates pain to some extent but fails in some tail classes and boundaries of objects. In the last column, we propose a discriminative per-pixel representation learning scheme for fine-grained class alignment to achieve better performance. It encourages the representation of each pixel in feature space to be close to the corresponding semantic prototype for both source and target domains, while being away from other prototypes.}\vspace{-4mm}
    \label{Fig_motivation}
\end{figure*}

Semantic segmentation aims to assign a semantic label to each pixel of an image. This task can be of paramount importance in various real-world scenarios (e.g., robot control~\cite{robot_control2,robot_control1}, autonomous driving~\cite{geiger2012autonomous,zhou2020autonomous}, medical diagnosis~\cite{chaitanya2020medical,ronneberger2015UNet}, etc.). Recent achievements in this field have been driven by deep neural networks with massive annotations~\cite{chen2018encoderdecoder,ZhenLZSFFQ20}. However, assembling such large-scale datasets with pixel-level annotations is onerous and even infeasible~\cite{Cordts2016Cityscapes}. An appealing alternative is to use images with dense ground-truth annotations from a simulator~\cite{stephan2016gtav,ros2016synthia}. Unfortunately, models purely trained from synthetic data usually undergo a noticeable performance drop when directly applied to real images due to the \textit{domain shift}~\cite{dataset_shift_in_ML09}. 
Therefore, domain adaptation (DA)~\cite{david2010theory,gong2012geodesic,lv2021pareto,ganin2015dann,YueZZ0DKS21,pan2010survey,pan2019transferrable,tsai2019domain,DICD,xie2021active,zou2019confidence,TSA} has become a promising direction, whose purpose is to facilitate the knowledge learned from a well-labeled source domain better transferring to another unlabeled target domain. 

The key to the domain adaptive semantic segmentation is that the divergence across domains should be reduced to lower the upper bound of error on the target domain. 
In the literature, most previous works exploit adversarial training~\cite{goodfellow2014gan} for distribution alignment in the input~\cite{li2019bidirectional,CycleGAN2017}, intermediate feature~\cite{hoffman2016fcns}, or output~\cite{tsai2018learning,liu2021bapa} space. However, existing techniques mostly strive to align marginal distribution in the domain level. In consequence, it is possible that decision boundaries traverse high-density regions of the target domain, rendering the learned classifier are less discriminative. To improve upon such alignment, latest works~\cite{chen2017crosscity,du2019ssf-dan,long2018conditional,luo2019taking,wang2020class,xie2018learning,ZhenWZLSSFQ20,zhang2019category} have shown that class conditional distributions should also be aligned, which plays a vital role in reducing the domain discrepancy. Nonetheless, since the conditional distributions are unknown on the target domain, the extracted semantic features could be noisy. 

As a matter of fact, the local semantic affiliations of each pixel in the feature space are crucial during adaptation. As depicted in Fig.~\ref{Fig_motivation}, we note that 1) Global alignment methods, e.g., AdaptSegNet~\cite{tsai2018learning}, are inclined to consider the domain divergence globally and ignore the underlying structures among classes, leading to domain-invariant but class-indistinguishable segmentation results. 2) Although some class alignment methods, e.g., CAG-UDA~\cite{zhang2019category}, achieve potential gains, they might remain inconsistent concerning the boundaries of objects or some tail classes (e.g., light, sign, bike). Such alignment does not incorporate abundant pixel-wise context and structural information, which usually obtain coarse-grained results. 3) Instead, our method explicitly investigates per-pixel discriminative representation learning and performs fine-grained class alignment guided by semantic prototypes, achieving more discriminative features and consistent performance.

In this paper, we provide a new perspective for tackling domain adaptation in semantic segmentation. The primary aim is to enhance the intra-class compactness and inter-class separability of the source domain (with annotations) in the pixel wise and transfer such discriminative information into the target domain (without annotations) via semantic prototype-based contrastive learning. Specifically, we first construct the representative semantic prototype for each individual class via source ground-truth labels, which can provide supervisory signals for learning discriminative pixel representations across the two domains. To ensure the reliability of the generated semantic prototypes, we dynamically update and distinguish different prototypes. Simultaneously, separation of pixel representations from different classes in source-domain data can be naturally guaranteed. Furthermore, for the target-domain data, representation of each pixel can be correspondingly divided into subsets according to its reliable pseudo label. In this way, for each pixel representation from both domains, we can properly construct one positive pixel-prototype pair and $C-1$ ($C$ is the number of semantic labels) negative pixel-prototype pairs. Finally, a novel contrastive loss is introduced to improve the discrimination of pixel representations, enforcing the positive-concentrated and negative-separated properties. 
In summary, we make the following contributions:
\begin{itemize}
    \item We propose a simple DA method for semantic segmentation that explicitly enhances pixel-wise intra-class compactness and pixel-wise inter-class separability.
    \item Directly applying contrastive learning techniques validated in image classification renders particular challenges in segmentation. We investigate a novel semantic prototype-based contrastive learning to effectively align pixel-wise representations with contextual semantic prototypes across domains.
    \item We conduct extensive experimental studies on four popular datasets including GTA5, Synscapes, SYNTHIA, and Cityscapes. Comprehensive analyses are conducted to validate the effectiveness of our method.
\end{itemize}

\section{Related work}

\subsection{Contrastive learning} Contrastive learning has become a dominant part in unsupervised learning~\cite{chen2020contrastive,he2020momentum,park2020contrastive,oord2018infoNCE,radford2021learning}. The intuition is that different augmented versions of an image should have similar representations and these representations should also differ from those of a distinct image. In contrast, we aim to learn pixel-wise representations to distinguish different areas in an image for segmentation task instead of facilitating learning of meaningful image-wise representations for classification task.  In essence, the optimization objective is different.

Recent works~\cite{alonso2021semi,xie2020contrastive_dense,wang2020contrastive_dense,kim2021learning,wenguan_exploring_ICCV} also generalize contrastive learning to pixel-level dense prediction tasks. However, these methods engage in the unsupervised pre-training~\cite{xie2020contrastive_dense}, the fully supervised setting~\cite{wenguan_exploring_ICCV} or semi-supervised setting~\cite{alonso2021semi}. And the recipes for semantic segmentation under the context of domain shift are yet to be built. To this end, we tailor multiple similar/dissimilar pixel-prototype pairs according to semantic prototypes for aligning source and target features. 

\subsection{Domain adaptive semantic segmentation} Generally, domain adaptation (DA) has been extensively explored to narrow the distribution mismatch between training and testing dataset for the image classification task~\cite{ganin2015dann,long2015dan,long2018conditional,peng2019moment,tzeng2015simultaneous,xie2018learning,JADA,GDCAN}. Not until recently has limited effort been made for semantic segmentation. Hoffman et al.~\cite{hoffman2016fcns} are the first to introduce DA to segmentation, where they consider feature alignment with additional category constraints. Enormous adversarial learning variants are proposed to learn domain-invariant features and they can be categorized into  input~\cite{Hoffman_cycada2017,li2019bidirectional}, feature~\cite{du2019ssf-dan,wang2020differential}, output~\cite{luo2019taking,tsai2018learning,yang2020adversarial} or patch~\cite{tsai2019domain} space adaptations. To name a few, Tsai et al.~\cite{tsai2018learning} consider segmentation as structured outputs, where images from different domains share strong similarities in semantic layout. Similarly, Luo et al.~\cite{luo2019taking} suggest applying different adversarial weights to different pixels. Wang et al..~\cite{wang2020class} incorporate class information into the discriminator to align features at a fine-grained level. Most recently, Kang et al.~\cite{kang2020pixel} and Melas-Kyriazi et al.~\cite{pixmatch2021_CVPR} provide a non-adversarial perspective to diminish domain gap via exploiting the correlations of pixels. However, implicit class alignment may be affected by class imbalance. On the contrary, we delve into considering per-pixel discriminative representation learning with the aid of semantic prototypes.

Another direction resorts to self-supervision~\cite{wang2021uncertainty,ProDA_2021_CVPR,lian2019pycda,pan2020unsupervised,Zhang_2017_ICCV,zou2018unsupervised,zou2019confidence,wang2021domain} to boost the segmentation performance, where the confident predictions of the unlabeled target data are used to fine-tune the model trained on the source domain. Zou et al.~\cite{zou2018unsupervised} first propose an iterative learning strategy with class balance and spatial prior in the target domain. In addition, a soft-assignment version of the pseudo label is proposed in~\cite{vu2019advent} to focus on ``most-confused'' pixels, which obtains better performance. Very recently, Pan et al.~\cite{pan2020unsupervised} propose a two-step self-supervised domain adaptation technique, which considers adapting from the easy image to the hard image within the target domain. However, they all rely on a good initialization and a difficult fine-tuning process. 

Of particular relevance to our work is the method CAG-UDA~\cite{zhang2019category}, which enforces category-aware feature alignment by minimizing distance between pixel feature and the corresponding category centroid. 
However, except for the semantic information, per-pixel feature also involves abundant structural information. Naively minimizing the distance loss only independently adapt semantic features across domains and thus is less discriminative. 
In contrast, we set forth a simple semantic prototype-based contrastive loss to learn discriminative pixel representations, which allows us to model pixel-wise intra-class compactness and pixel-wise inter-class separability across domains. 
In addition, the prototypes provided by our method are gradually updated and refined, while CAG-UDA only supports fixed targets during training. Note that our updating strategy (will be discuss in Section~\ref{sec:semantic_prototypes}) dynamically accommodates the parameters of the network.
Such alignment strengthens the connections between pixel representations of both domains and the corresponding prototypes, which enables the adaptation in a more accurate and stable manner.

\section{Method}\label{sec:method}

\subsection{Overview of the framework}
In domain adaptation, given a collection of images and ground-truth labels from the source domain denoted as $\Dm_s=\{X_{s_i}, Y_{s_i}\}_{i=1}^{n_s}$, as well as unlabeled images from the target domain $\Dm_t=\{X_{t_j}\}_{j=1}^{n_t}$, the problem is to adapt a segmentation model from source domain $\Dm_s$ to target domain $\Dm_t$. Here $n_s$ and $n_t$ are the numbers of samples from two domains. Note that $X_s, X_t \in \Rb^{H\times W\times 3}$, and $Y_s \in \mathbb{B}^{H\times W\times C}$ with pixel-level one-hot vectors, where $H$ and $W$ represent the spatial dimensions of the image and $C$ is the number of semantic class labels. 

\begin{figure*}[t]
    \centering
    \includegraphics[width=\textwidth]{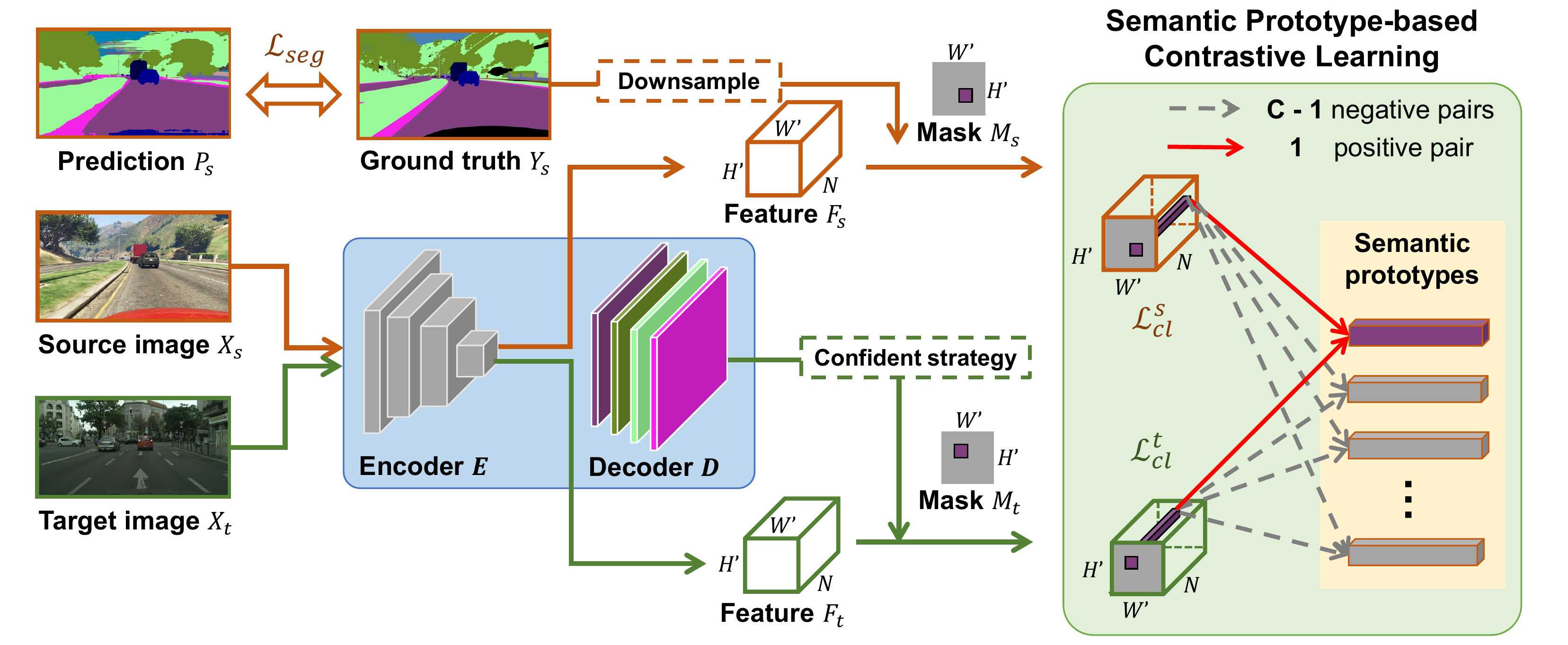}
    \caption{{\bf Overview of the proposed method}. Images in the source (\textcolor[RGB]{197, 90, 17}{brown} arrows) and target (\textcolor[RGB]{84, 193, 53}{green} arrows) domains are randomly selected and passed through the segmentation network ($E$ and $D$) to get final predictions. For the source data, a segmentation loss $\Lm_{seg}$ is computed based on the ground truth $Y_s$. We separate each pixel representation obtained from $E$ in both domains according to their masks $M_{s}$ and $M_t$ and pass them to semantic prototype-based contrastive learning module. As a result, clusters of pixel representations belonging to the same prototype are pulled together in the feature space while simultaneously pushing apart from other prototypes, which improves pixel-wise compactness and separability.}
    \label{Fig_framework}
\end{figure*}

The overall framework is depicted in Figure~\ref{Fig_framework}.
Specifically, the segmentation network consists of an encoder $E$ and a decoder (multi-class classifier) $D$. 
We first pass both source and target images through the encoder $E$ and obtain their feature maps $F_s\,,F_t \in \Rb^{H'\times W'\times N}$, then pass the features to the decoder $D$ and get the predictions $O_s\,,O_t \in \Rb^{H'\times W' \times C}$. Finally, we acquire the pixel-wise output predictions $P_s, P_t \in \Rb^{H\times W\times C}$ after the upsample and softmax operations.

The key in our work is to ensure representation of each pixel in feature maps $F_s$ and $F_t$ to be close to their corresponding source semantic prototype, while being pushed away from other semantic prototypes. In this way, features from the same class in the two domains tend to be clustered compactly, effectively boosting the model generalization capability. 
To achieve this, source class centroids can be calculated as representative semantic prototypes. Then each pixel representation in feature space derived from encoder $E$ will be considered and processed separately given their predicted masks $M_s$ and $M_t$. Based on this, an effective semantic prototype-based contrastive loss is proposed to align conditional distributions across domains via learning discriminative representations of pixels. Note that the proposed contrastive loss can be used in both source and target domains simultaneously. For one thing, when the loss is applied in $F_s$, the encoder is able to yield more discriminative features for decoder, which could increase the robustness of semantic segmentation model. For another, the designed pixel-wise contrastive loss for target feature map $F_t$ facilitates transferring knowledge from source to target explicitly, achieving more precise alignment.

\subsection{Semantic prototype-based contrastive learning}
\paragraph{Contrastive learning revisit} 
Contrastive learning and its variants~\cite{hadsell2006dimensionality,he2020momentum,oord2018infoNCE} aim to learn representations from unlabeled data organized in similar/dissimilar pairs, in which meaningful and discriminative representations are learned through a contrastive loss function $\ell$. If we denote the query and key vectors as $u, v\in \Rb^N$, where $N$ is the dimension of embedding space, the loss $\ell$ is designed to reflect the incompatibility of data pair $(u,v)$. Here, $v^+$ and $\{v^-\}$ represent the positive key and negative key sets with respect to query $u$, respectively. In a word, a desirable $\ell$ returns low value given the positive pair $(u,v^+)$, but achieves high loss for the negative pairs $(u,\{v^-\})$, which could effectively force positive query and key vectors to be similar while distinct from negative ones. 

Inspired by~\cite{oord2018infoNCE}, a standard contrastive learning problem can be treated as a ``two-class'' classification problem, and the loss for query $u$ is formulated as:
 \begin{small}
    \begin{align}
        \ell({u,v^+,\{v^-\}}) = -\log \frac{\exp\left(u{\cdot}v^+ / \tau\right)}{\exp\left(u{\cdot}v^+ / \tau\right) + {\displaystyle\sum_{v^-} \exp\left(u{\cdot}v^- / \tau\right)}}.
    \end{align}
\end{small}%
As shown above, the contrastive loss is essentially based on the softmax formulation with a temperature $\tau$~\cite{ge2020contrastive,park2020contrastive,oord2018infoNCE} to cluster the positive pair $(u,v^+)$ as well as push away the negative pairs $(u,\{v^-\})$. Hence, the meaningful representations can be learned via multiple unlabeled data. 

In essence, expect for the formulation design of contrastive loss, how to construct the positive and negative key-query pairs remains challenging for the successful application of contrastive learning. In this paper, we novelly set semantic prototypes for all classes as key vectors, and pixel representations in the intermediate layers of both domains as query vectors. Meanwhile, combined with the proposed contrastive learning loss, the cross-domain discriminative clustering structures will be learned for better adaptation.

\subsubsection{Semantic prototype learning}
\label{sec:semantic_prototypes}
In order to encode all pixel representations from both domains for discriminative feature learning and knowledge transfer, for each class, we intend to cluster the representations around their corresponding class centroid (semantic prototype). In the meantime, a dynamic semantic prototypes updating strategy is proposed during the training.

\paragraph{Prototype initialization.} The initial semantic prototypes can be calculated as the mean feature representation of each class from source domain since the ground-truth semantic labels are available and can naturally provide valuable supervisions. To assign each source pixel feature with a reliable semantic label, we can directly leverage the downsampled ground-truth label map $\bar{Y}_s$ (a downsampled version of ${Y}_s$) to obtain a mask $M_s$ for an input image by $M_s^{(h',w')} = \mathop{\arg\max}\limits_{c}  \bar{Y}_s^{(h',w',c)}\,.$
Based on the mask $M_s$, the semantic prototype of $c$-th class on the entire source domain can be calculated as:
 \begin{small}
\begin{align}
    \mu^c = \frac{1}{n_s}\sum_i \frac{1}{|\Phi_{i}^c|} \sum_{h'\,, w'} \mathbb{I}_{[M_{s_i}^{(h',w')}=c]} F_{s_i}^{(h'\,, w')}\,,
    \label{eq:semantic_prototype_initialize}
\end{align}
\end{small}%
where $\mathbb{I}$ is an indicator function which returns 1 if the condition holds or 0 otherwise. $\Phi_{i}^c$ denotes the pixel set that contains all the pixel features belonging to the $c$-th semantic class within $F_{s_i}$ and $|\cdot|$ is the number of pixels in the set. Note that the semantic prototypes are only initialized once at the beginning of the learning algorithm by performing the forward computation, and the prototypes can be continuously updated as the learning proceeds. 

\paragraph{Prototype updating.} At each iteration, the encoded pixel features of each image are different, and should be involved in the prototype updating process to represent the latest semantic knowledge. Technically, the $c$-th centroid $\mu^c$ is updated with the mean of the encoded pixel features belonging to class $c$ within a new feature map $F_{s}$ as follows:%
 \begin{small}
\end{small}%
 \begin{small}
    \begin{align}
        \mu^c \leftarrow \alpha \mu^c + \frac{1 - \alpha}{|\Phi^c|} \sum_{h'\,, w'} \mathbb{I}_{[M_s^{(h',w')}=c]} F_{s}^{(h'\,, w')}\,.
        \label{eq:semantic_prototype_update}
    \end{align}
\end{small}%
Here $\alpha \in [0,1]$ is a coefficient for updating semantic prototypes, and $\alpha$ is empirically set as 0.1 in this paper. Note that $\alpha$ = 1.0 denotes that the prototypes are fixed during the training process. With the updated prototypes, our method can dynamically guide the cross-domain pixel representations to cluster around the corresponding prototype, and enforce the category alignment across domains.

\subsubsection{Fine-grained class alignment}
\label{sec:alignment}
Recently, several prior methods~\cite{du2019ssf-dan,luo2019taking,wang2020differential,zhang2019category,pixmatch2021_CVPR} have leveraged class alignment to remedy the domain gap. However, most of them are coarse-grained adaptation and often neglect the discriminative knowledge of pixel representations, which severely limits their potential capability in pixel-level prediction tasks. By contrast, in our work, we explicitly model the intra-class compactness and inter-class separability across domains in the pixel wise by proposing a semantic prototype-based contrastive learning framework.

Specifically, to design an effective contrastive loss for domain adaptive semantic segmentation, the key is to construct the appropriate query-key pairs. Different from the conventional contrastive learning methods, which rely on the data itself for supervision, we believe that the source semantic prototypes can provide valuable guidance for the pixel samples in both domains. In essence, the proposed contrastive loss is to force the features from the same class to concentrate together while pushing away different clusters for both domains, enhancing the model robustness and improving cross-domain knowledge transfer. 

Therefore, if we define each pixel obtained from the encoder $E$ as a query vector, the corresponding semantic prototype from the same class is reasonably set as the unique positive key vector, and semantic prototypes in other $C-1$ classes should be negative keys. In practice, we normalize these features onto a unit sphere to prevent the space from collapsing or expanding. Then in source domain, by integrating the source mask $M_s$ and feature map $F_{s}$, the proposed contrastive loss on source domain is
 \begin{small}
    \begin{align}
        \Lm_{cl}^s = \sum_i\sum_{h',w'} \sum_{c} \mathbb{I}_{[M_{s_i}^{(h',w')}=c]} \ell({F_{s_i}^{(h',w')}\,,\mu^{c}\,,\{\mu^{c-}\}})\,,
        \label{eq:source_local_loss}
    \end{align}
\end{small}%
where $\mu^{c}$ and $\{\mu^{c-}\}$ represent the corresponding positive and $C-1$ negative semantic prototypes, respectively.
Optimizing $\Lm_{cl}^s$ can leverage label information effectively in the feature space. In addition to this, we conduct the proposed contrastive learning for all target pixel features as well to transfer the supervised knowledge from source to target.

However, for target domain data, training error could be amplified by noisy predictions when generating the target mask. To remedy this, we employ a confident strategy with a confidence threshold $\sigma_o^c$ for each class $c$ individually. In detail, firstly, a confidence map is generated according to the segmentation prediction map $O_t$, where the confidence value is the maximum item of the softmax output in each pixel. This enables the pseudo label at each pixel to be associated with a confidence value, i.e., the prediction probability. Secondly, if the median confidence value for a certain class is above 0.9, then the confidence threshold for that class is set to 0.9; otherwise it is set to the median confidence value. With the $\sigma_o^c$ being set, we can define target mask in the feature space as follows:
 \begin{small}
    \begin{align}
        M_t^{(h',w')} = \mathop{\arg\max}\limits_{c} \mathbb{I}_{[O_t^{(h',w',c)}>\sigma_o^c]} O_t^{(h',w',c)}\,.
        \label{eq:target_mask}
    \end{align}
\end{small}%

Similarly, we can obtain target contrastive loss for all the target pixel representations as:
 \begin{small}
    \begin{align}
        \Lm_{cl}^t = \sum_j\sum_{h',w'} \sum_{c} \mathbb{I}_{[M_{t_j}^{(h',w')}=c]} \ell({F_{t_j} ^{(h',w')}\,,\mu^{c}\,,\{\mu^{c-}\}})\,.
        \label{eq:target_local_loss}
    \end{align}
\end{small}%
By observing Eq. \eqref{eq:target_local_loss}, optimizing $\Lm_{cl}^t$ can not only pull each target pixel sample to its corresponding semantic prototype and achieve class-level knowledge transfer, but also preserve the target intrinsic discriminative structure, which will boost the final model generalization capability.

\subsection{Objective}
In this work, we follow a popular two-step training procedure~\cite{du2019ssf-dan,pan2020unsupervised,wang2020class,wang2020differential,zhang2019category} to reduce the domain shift and improve the performance of the segmentation model on the target domain. First, given source images with ground-truth labels, we employ the broadly used cross-entropy loss to guarantee a small source error,
 \begin{small}
    \begin{align}
        \Lm_{seg} = -\sum_i \sum_{h,w} \sum_{c} Y_{s_i}^{(h,w,c)} \log\left(P_{s_i}^{(h,w,c)}\right)\,.
        \label{eq:segmentation_loss}
    \end{align} 
\end{small}%
Combining $\Lm_{seg}\,,\Lm_{cl}^s\,,\Lm_{cl}^t$ with a balancing weight $\lambda$, we are able to close the domain gap between the source and target data and perform the segmentation task, and the overall objective is formulated as follows:
 \begin{small}
\begin{align}
    \mathop{\min}\limits_{E\,,D} \Lm_{seg} + \lambda (\Lm_{cl}^s + \Lm_{cl}^t).
    \label{eq:overall_loss}
\end{align}
\end{small}%
By optimizing Eq.~\eqref{eq:overall_loss}, clusters of pixels belonging to the same class are pulled together in the feature space while synchronously pushing apart from other classes. In this way, our method can simultaneously minimize the domain gap across domains as well as enhance the intra-class compactness and inter-class separability in a unified framework.

Second, once the alignment is finished, we can generate the reliable pseudo labels for target data by choosing the confidence threshold $\sigma_p^c$ for each class respectively according to the output predictions $P_t$ (similar to setting $\sigma_o^c$). The target pseudo labels in the output space are obtained as follows: $\hat{Y}_t^{(h,w)} = \mathop{\arg\max}\limits_{c} \mathbb{I}_{[P_t^{(h,w,c)}>\sigma_p^c]} P_t^{(h, w,c)}\,.$
Then, we fine-tune the model via optimizing a self supervision loss $\Lm_{ssl}$ in Eq.~\eqref{eq:self_training} on the entire target training data to make the model more adaptive to the target domain.
 \begin{small}
    \begin{align}
        \mathop{\min}\limits_{E\,,D} -\sum_{j} \sum_{h,w} \sum_{c} \mathbb{I}_{[\hat{Y}_{t_j}^{(h,w)}=c]} \log\left(P_{t_j}^{(h,w,c)}\right) \,.
        \label{eq:self_training}
    \end{align}
\end{small}%

\section{Experiments}
\subsection{Setups}
\textbf{Datasets.} We evaluate our method under the ``Sim-to-Real'' scenario with four popular benchmark datasets, i.e., transferring from the synthetic images (GTA5~\cite{stephan2016gtav}, Synscapes~\cite{wrenninge2018synscapes}, and SYNTHIA~\cite{ros2016synthia}) to the real images (Cityscapes~\cite{Cordts2016Cityscapes}). {\bf Cityscapes} includes 5,000 urban scene images of resolution 2048$\times$1024. They are splitted into training, validation, and testing set with 2,975, 500, and 1,525 images respectively. Similar to~\cite{tsai2018learning,zou2019confidence}, we evaluate our adapted model on the validation set.
{\bf GTA5} contains 24,966 images with the resolution of 1914$\times$1052. {\bf Synscapes} contains 25,000 images with a resolution of 1,440$\times$720. {\bf SYNTHIA} offers 9,400 images of resolution 1280$\times$760. 

\textbf{Network architectures.} For fair comparison, we utilize the DeepLab-v2 framework~\cite{chen2018deeplab} with ResNet-101~\cite{he2016deep} or VGG-16~\cite{simonyan2015vgg} as the base encoder $E$. Note that some state-of-the-art methods~\cite{lian2019pycda,zhang2019category} use different backbone networks~\cite{chen2018encoderdecoder,zhao2017pspnet}, and we present them as references to analyze how much our method can improve. All models are pre-trained on ImageNet~\cite{deng2009imagenet}. To better capture the scene context, Atrous Spatial Pyramid Pooling (ASPP)~\cite{chen2018deeplab} is used as decoder $D$ and applied on the encoder's outputs. Following~\cite{luo2019taking,tsai2018learning,zou2019confidence}, sampling rates are fixed as \{6, 12, 18, 24\} and we modify the stride and dilation rate of the last layers to produce denser feature maps with larger field-of-views.

\textbf{Training details.} Our training is carried out on 4 Tesla V100 GPUs. And we implement all methods with PyTorch~\cite{paszke2019pytorch}. To train the segmentation network, we adopt the SGD optimizer where the momentum is 0.9 and the weight decay is $10^{-4}$. The learning rate is initially set to $2.5\times 10^{-4}$ and is decreased following a `poly' learning rate policy with power of 0.9. $\alpha$ is constantly set to 0.1 and $\lambda$ is set to 1.0 for all experiments. 
Regarding the training procedure, we first use the source data to train the network as well as to align the output distributions following~\cite{tsai2018learning} as the baseline. Then the network is fine-tuned using our method for 40k iterations with batch size of 8 (four are source images and the other four are target images). Some data augmentations (e.g., color jittering and random horizontal flip etc.) are used to prevent overfitting. Finally, similar to~\cite{li2019bidirectional,pan2020unsupervised,wang2020class,wang2020differential}, we apply the self supervision loss to further improve the performance on the target domain. 

\textbf{Evaluation metrics.} We employ the PSACAL VOC Intersection-over-Union (IoU) as the evaluation metric~\cite{everingham2015IoU}, i.e, IoU =$\frac{TP}{TP+FP+FN}$, where $TP$, $FP$, and $FN$ stand for the amount of true positive, false positive and false negative pixels, respectively, determined over the whole test set. For GTA5 $\rightarrow$ Cityscapes and Synscapes $\rightarrow$ Cityscapes tasks, we report the results on the common 19 classes and the tail classes. For SYNTHIA $\rightarrow$ Cityscapes task, we report the results over 16 and 13 classes.

\begin{table*}[t]
    \begin{center}
        \caption{Experimental results for GTA5 $\to$ Cityscapes. mIoU$_{tail}$ denotes the mean IoU of the tail classes in \textcolor{blue}{blue}.} \label{table:gta}
        \resizebox{\textwidth}{!}{
        \begin{tabular}{c|l c c c c c c c c c c c c c c c c c c c|c|c}
            \toprule[1.2pt]
            Backbone & Method  & \rotatebox{60}{road} & \rotatebox{60}{side.} & \rotatebox{60}{buil.} & \rotatebox{60}{\textcolor{blue}{wall}} & \rotatebox{60}{\textcolor{blue}{fence}} & \rotatebox{60}{pole} & \rotatebox{60}{\textcolor{blue}{light}} & \rotatebox{60}{\textcolor{blue}{sign}} & \rotatebox{60}{veg} & \rotatebox{60}{\textcolor{blue}{terr.}} & \rotatebox{60}{sky} & \rotatebox{60}{pers.} & \rotatebox{60}{\textcolor{blue}{rider}} & \rotatebox{60}{car} & \rotatebox{60}{\textcolor{blue}{truck}} & \rotatebox{60}{\textcolor{blue}{bus}} & \rotatebox{60}{\textcolor{blue}{train}} & \rotatebox{60}{\textcolor{blue}{mbike}} & \rotatebox{60}{\textcolor{blue}{bike}} & mIoU & mIoU$_{tail}$ \\
            \hline
            \multirow{8}{*}{VGG-16}
            &AdaptSegNet~\cite{tsai2018learning} & 87.3 & 29.8 & 78.6 & 21.1 & 18.2 & 22.5 & 21.5 & 11.0 & 79.7 & 29.6 & 71.3 & 46.8 & 6.5 & 80.1 & 23.0 & 26.9 & 0.0 & 10.6 & 0.3 & 35.0 & 15.3 \\
            &CBST~\cite{zou2018unsupervised} & 90.4 & 50.8 & 72.0 & 18.3 & 9.5 & 27.2 & 28.6 & 14.1 & 82.4 & 25.1 & 70.8 & 42.6 & 14.5 & 76.9 & 5.9 & 12.5 & 1.2 & 14.0 & \bf 28.6 & 36.1 & 15.7 \\
            & AdvEnt~\cite{vu2019advent} & 86.9 & 28.7 & 78.7 & 28.5 & 25.2 & 17.1 & 20.3 & 10.9 & 80.0 & 26.4 & 70.2 & 47.1 & 8.4 & 81.5 & 26.0 & 17.2 & \bf 18.9 & 11.7 & 1.6 & 36.1 & 17.7 \\
            & CLAN~\cite{luo2019taking} & 88.0 & 30.6 & 79.2 & 23.4 & 20.5 & 26.1 & 23.0 & 14.8 & 81.6 & 34.5 & 72.0 & 45.8 & 7.9 & 80.5 & 26.6 & 29.9 & 0.0 & 10.7 & 0.0 & 36.6 & 17.4 \\
            & APODA~\cite{YangXLQSLL20} & 88.4 & 34.2 & 77.6 & 23.7 & 18.3 & 24.8 & 24.9 & 12.4 & 80.7 & 30.4 & 68.6 & 48.9 & 17.9 & 80.8 & 27.0 & 27.2 & 6.2 & 19.1 & 10.2 & 38.0 & 16.7 \\
            & CrCDA~\cite{huang2020contextual} & 86.8 & 37.5 & 80.4& 30.7 & 18.1 & 26.8 & 25.3 & 15.1 & 81.5 & 30.9 & 72.1 & 52.8 & 19.0 & 82.1 & 25.4& 29.2 & 10.1 & 15.8 & 3.7 & 39.1 & 20.3 \\
            & FADA~\cite{wang2020class} & \bf 92.3 & \bf 51.1 & 83.7 & \bf 33.1 & \bf 29.1 & 28.5 & 28.0 & \bf 21.0 & 82.6 & 32.6 & \bf 85.3 & 55.2 & 28.8 & 83.5 & 24.4 & 37.4 & 0.0 & 21.1 & 15.2 & 43.8 & 24.6 \\
            & \cellcolor{Gray}Ours & \cellcolor{Gray} 91.3 & \cellcolor{Gray} 44.9 & \cellcolor{Gray} \bf 84.0 & \cellcolor{Gray} 31.9 & \cellcolor{Gray} 27.3 & \cellcolor{Gray}\bf 35.4 & \cellcolor{Gray}\bf 36.3 & \cellcolor{Gray} 15.7 & \cellcolor{Gray} \bf 83.6 & \cellcolor{Gray} \bf 34.7 & \cellcolor{Gray} 82.1 & \cellcolor{Gray} \bf 58.2 & \cellcolor{Gray}\bf 28.9 & \cellcolor{Gray}\bf 85.7 & \cellcolor{Gray}\bf 29.0 & \cellcolor{Gray}\bf 38.2 & \cellcolor{Gray} 0.0 & \cellcolor{Gray}\bf 24.7 & \cellcolor{Gray} 19.5 & \cellcolor{Gray}\bf 44.8 & \cellcolor{Gray}\bf 26.0 \\
            \hline
            \hline
            \multirow{22}{*}{ResNet-101}
            &Source Only & 65.0 & 16.1 & 68.7 & 18.6 & 16.8 & 21.3 & 31.4 & 11.2 & 83.0 & 22.0 & 78.0 & 54.4 & 33.8 & 73.9 & 12.7 & 30.7 & 13.7 &28.1 & 19.7 & 36.8 & 21.7 \\
            &AdaptSegNet~\cite{tsai2018learning} & 86.5 & 36.0 & 79.9 & 23.4 & 23.3 & 23.9 & 35.2 & 14.8 & 83.4 & 33.3 & 75.6 & 58.5 & 27.6 & 73.7 & 32.5 & 35.4 & 3.9 & 30.1 & 28.1 & 42.4 & 26.1 \\
            &CLAN~\cite{luo2019taking} & 87.0 & 27.1 & 79.6 & 27.3 & 23.3 &28.3 & 35.5 & 24.2 & 83.6 & 27.4 & 74.2 & 58.6 & 28.0 & 76.2 &   33.1 & 36.7 & 6.7 & 31.9 & 31.4 & 43.2 & 27.8 \\
            &AdvEnt~\cite{vu2019advent} & 89.9 & 36.5 & 81.6 & 29.2 & 25.2 & 28.5 & 32.3 & 22.4 & 83.9 & 34.0 & 77.1 & 57.4 & 27.9 & 83.7 & 29.4 & 39.1 & 1.5 & 28.4 & 23.3 & 43.8 & 26.6 \\
            &SSF-DAN~\cite{du2019ssf-dan} & 90.3 & 38.9 & 81.7 & 24.8 & 22.9 & 30.5 & 37.0 & 21.2 & 84.8 & 38.8 & 76.9 & 58.8 & 30.7 & 85.7 & 30.6 & 38.1 & 5.9 & 28.3 & 36.9 & 45.4 & 28.7 \\
            &CBST~\cite{zou2018unsupervised} & 91.8 & 53.5 & 80.5 & 32.7 & 21.0 & 34.0 & 28.9 & 20.4 & 83.9 & 34.2 & 80.9 & 53.1 & 24.0 & 82.7 & 30.3 & 35.9 & 16.0 & 25.9 & 42.8 & 45.9 & 28.4 \\   
            & APODA~\cite{YangXLQSLL20} & 85.6 & 32.8 & 79.0 & 29.5 & 25.5 & 26.8 & 34.6 & 19.9 & 83.7 & 40.6 & 77.9 & 59.2 & 28.3 & 84.6 & 34.6 & 49.2 & 8.0 & 32.6 & 39.6 & 45.9 & 31.1 \\         
            &IntraDA~\cite{pan2020unsupervised} &  90.6 & 37.1 & 82.6 & 30.1 & 19.1 & 29.5 & 32.4 & 20.6 & 85.7 & 40.5 & 79.7 & 58.7 & 31.1 & 86.3 & 31.5 & 48.3 & 0.0 & 30.2 & 35.8 & 46.3 & 29.1 \\ 
            &CRST~\cite{zou2019confidence} & 91.0 & 55.4 & 80.0 & 33.7 & 21.4 & 37.3 & 32.9 & 24.5 & 85.0 & 34.1 & 80.8 & 57.7 & 24.6 & 84.1 & 27.8 & 30.1 & 26.9 & 26.0 & 42.3 & 47.1 & 29.5 \\
            &PyCDA~\cite{lian2019pycda} & 90.5 & 36.3 & 84.4 & 32.4 & 28.7 & 34.6 & 36.4 & 31.5 & \textbf{86.8} & 37.9 & 78.5 & 62.3 & 21.5 & 85.6 & 27.9 & 34.8 & 18.0 & 22.9 & \textbf{49.3} & 47.4 & 31.0 \\
            &PLCA~\cite{kang2020pixel} & 84.0 & 30.4 & 82.4 & 35.3 & 24.8 & 32.2 & 36.8 & 24.5 & 85.5 & 37.2 & 78.6 & \bf 66.9 & 32.8 & 85.5 & 40.4 & 48.0 & 8.8 & 29.8 & 41.8 & 47.7 & 32.7\\
            & WeakSegDA~\cite{Paul_WeakSegDA_ECCV20} & 91.6 & 47.4 & 84.0 & 30.4 & 28.3 & 31.4 & 37.4 & 35.4 & 83.9 & 38.3 & 83.9 & 61.2 & 28.2 & 83.7 & 28.8 & 41.3 & 8.8 & 24.7 & 46.4 & 48.2 & 31.6 \\
            &BDL~\cite{li2019bidirectional} & 91.0 & 44.7 & 84.2 & 34.6 & 27.6 & 30.2 & 36.0 & 36.0 & 85.0 & 43.6 & 83.0 & 58.6 & 31.6 & 83.3 & 35.3 & 49.7 & 3.3 & 28.8 & 35.6 & 48.5 & 32.9 \\
            & CrCDA~\cite{huang2020contextual} & 92.4 & 55.3 & 82.3 & 31.2 & 29.1 & 32.5 & 33.2 & 35.6 & 83.5 & 34.8 & 84.2 & 58.9 & 32.2 & 84.7& 40.6 & 46.1 & 2.1 & 31.1 & 32.7 & 48.6 & 31.7 \\
            &SIM~\cite{wang2020differential} & 90.6 & 44.7 & 84.8 & 34.3 & 28.7 & 31.6 & 35.0 & 37.6 & 84.7 & 43.3 & 85.3 & 57.0 & 31.5 & 83.8 & 42.6 & 48.5 & 1.9 & 30.4 & 39.0 & 49.2 & 33.9 \\
            &FADA~\cite{wang2020class} & 92.5 & 47.5 & 85.1 & 37.6 & \bf 32.8 & 33.4 & 33.8 & 18.4 & 85.3 & 37.7 & 83.5 & 63.2 & \bf 39.7 & 87.5 & 32.9 & 47.8 & 1.6 & 34.9 & 39.5 & 49.2 & 32.4 \\  
            & LDR~\cite{yang2020label-driven} & 90.8 & 41.4 & 84.7 &  35.1 &27.5&31.2&38.0&32.8&85.6&42.1&84.9&59.6&34.4&85.0&42.8&52.7&3.4&30.9&38.1&49.5 & 34.3 \\
            & CCM~\cite{li2020content} & \bf 93.5 & \bf 57.6 & 84.6 & 39.3 & 24.1 & 25.2 & 35.0 & 17.3 & 85.0 & 40.6 & 86.5 & 58.7 & 28.7 & 85.8 & \bf 49.0 & \bf 56.4 & 5.4 & 31.9 & 43.2 & 49.9 & 33.7 \\
            &CAG-UDA~\cite{zhang2019category} & 90.4 & 51.6 & 83.8 & 34.2 & 27.8 & \bf 38.4 & 25.3 & \bf 48.4 & 85.4 & 38.2 & 78.1 & 58.6 & 34.6 & 84.7 & 21.9 & 42.7 & \bf 41.1 & 29.3 & 37.2 & 50.2 & 34.6 \\
            & PixelMatch~\cite{pixmatch2021_CVPR} & 91.6 & 51.2 & 84.7 & 37.3 & 29.1 & 24.6 & 31.3 & 37.2 & 86.5 & 44.3 & 85.3 & 62.8 & 22.6 & 87.6 & 38.9 & 52.3 & 0.7 & 37.2 & 50.0 & 50.3 & 34.6 \\
            & FDA~\cite{yang2020fda} & 92.5 & 53.3 & 82.4 & 26.5 & 27.6 & 36.4 & 40.6 & 38.9 & 82.3 & 39.8 & 78.0 & 62.6 & 34.4 & 84.9 & 34.1 & 53.1 &  16.9 & 27.7 & 46.4 &  50.5 & 35.1 \\
            &\cellcolor{Gray}Ours & \cellcolor{Gray}90.3 & \cellcolor{Gray}50.3 & \cellcolor{Gray}\bf 85.7 & \cellcolor{Gray}\bf 45.3 & \cellcolor{Gray}28.4 & \cellcolor{Gray}36.8 & \cellcolor{Gray}\bf 42.2 & \cellcolor{Gray}22.3 & \cellcolor{Gray}85.1 & \cellcolor{Gray}\bf 43.6 & \cellcolor{Gray}\bf 87.2 & \cellcolor{Gray}62.8 & \cellcolor{Gray}39.0 & \cellcolor{Gray}\bf 87.8 & \cellcolor{Gray}41.3 & \cellcolor{Gray}53.9 & \cellcolor{Gray}17.7 & \cellcolor{Gray}\bf 35.9 & \cellcolor{Gray}33.8 & \cellcolor{Gray}\bf 52.1 & \cellcolor{Gray}\bf 36.7 \\
            \bottomrule[1.2pt]
        \end{tabular}
        }
        \vspace{-4mm}
    \end{center}
\end{table*}

\begin{table*}[t]
    \begin{center}
        \caption{Experimental results for Synscapes $\to$ Cityscapes.} \label{table:synscapes}
        \resizebox{\textwidth}{!}{
            \begin{tabular}{c|l c c c c c c c c c c c c c c c c c c c|c|c}
            \toprule[1.2pt]
            Backbone & Method  & \rotatebox{60}{road} & \rotatebox{60}{side.} & \rotatebox{60}{buil.} & \rotatebox{60}{\textcolor{blue}{wall}} & \rotatebox{60}{\textcolor{blue}{fence}} & \rotatebox{60}{pole} & \rotatebox{60}{\textcolor{blue}{light}} & \rotatebox{60}{\textcolor{blue}{sign}} & \rotatebox{60}{veg} & \rotatebox{60}{\textcolor{blue}{terr.}} & \rotatebox{60}{sky} & \rotatebox{60}{pers.} & \rotatebox{60}{\textcolor{blue}{rider}} & \rotatebox{60}{car} & \rotatebox{60}{\textcolor{blue}{truck}} & \rotatebox{60}{\textcolor{blue}{bus}} & \rotatebox{60}{\textcolor{blue}{train}} & \rotatebox{60}{\textcolor{blue}{mbike}} & \rotatebox{60}{\textcolor{blue}{bike}} & mIoU & mIoU$_{tail}$ \\
            \hline
            \multirow{4}{*}{ResNet-101}
            &Source Only  & 81.8 & 40.6 & 76.1 & 23.3 & 16.8 & 36.9 & 36.8 & 40.1 & 83.0 & 34.8 & 84.9 & 59.9 & 37.7 & 78.4 & 20.4 & 20.5 & 7.8 & 27.3 & 52.5 & 45.3 & 28.9 \\
            &AdaptSegNet~\cite{tsai2018learning} & \bf 94.2 & \bf 60.9 & 85.1 & 29.1 & 25.2 & 38.6 & 43.9 & 40.8 & 85.2 & 29.7 & \bf 88.2 & 64.4 & 40.6 & 85.8 & 31.5 & 43.0 & 28.3 & 30.5 & 56.7 & 52.7 & 36.3 \\
            &IntraDA~\cite{pan2020unsupervised} & 94.0 & 60.0 & 84.9 & 29.5 & 26.2 & 38.5 & 41.6 & 43.7 & 85.3 & 31.7 & \bf 88.2 & 66.3 & 44.7 & 85.7 & 30.7 &  \textbf{53.0} & 29.5 & 36.5 & 60.2 & 54.2 & 38.8 \\
            &\cellcolor{Gray} Ours & \cellcolor{Gray}92.7 & \cellcolor{Gray}49.4 & \cellcolor{Gray}\bf 86.7 & \cellcolor{Gray}\bf 37.2 & \cellcolor{Gray}\bf 38.9 & \cellcolor{Gray}\bf 40.0 & \cellcolor{Gray}\bf 48.5 & \cellcolor{Gray}\bf 45.8 & \cellcolor{Gray}\bf 87.4 & \cellcolor{Gray}\bf 41.8 & \cellcolor{Gray}\bf 88.2 & \cellcolor{Gray}\bf 69.0 & \cellcolor{Gray}\bf 45.3 & \cellcolor{Gray}\bf 86.6 & \cellcolor{Gray}\bf 33.7 & \cellcolor{Gray}51.2 & \cellcolor{Gray}\bf 38.1 & \cellcolor{Gray}\bf 43.6 & \cellcolor{Gray}\bf 61.0 & \cellcolor{Gray}\bf 57.1 & \cellcolor{Gray}\bf 44.1 \\
            \bottomrule[1.2pt]
            \end{tabular}
        }
        \vspace{-5mm}
    \end{center}
\end{table*}

\subsection{Comparison with state-of-the-art methods}
We compare our method with previous state-of-the-arts in Table~\ref{table:gta}, Table~\ref{table:synscapes}, and Table~\ref{table:synthetic}. All the models utilize DeepLab-v2~\cite{chen2018deeplab} framework, except that PyCDA is based on PSPNet~\cite{zhao2017pspnet} and CAG-UDA is based on DeepLab-v3+\footnote{\url{https://github.com/RogerZhangzz/CAG_UDA/issues/6}}~\cite{chen2018encoderdecoder}. It can be seen that our method outperforms all the existing competing methods and achieves new state-of-the-art performance in terms of mIoU. 

\begin{figure*}
    \centering
    \includegraphics[width=\textwidth]{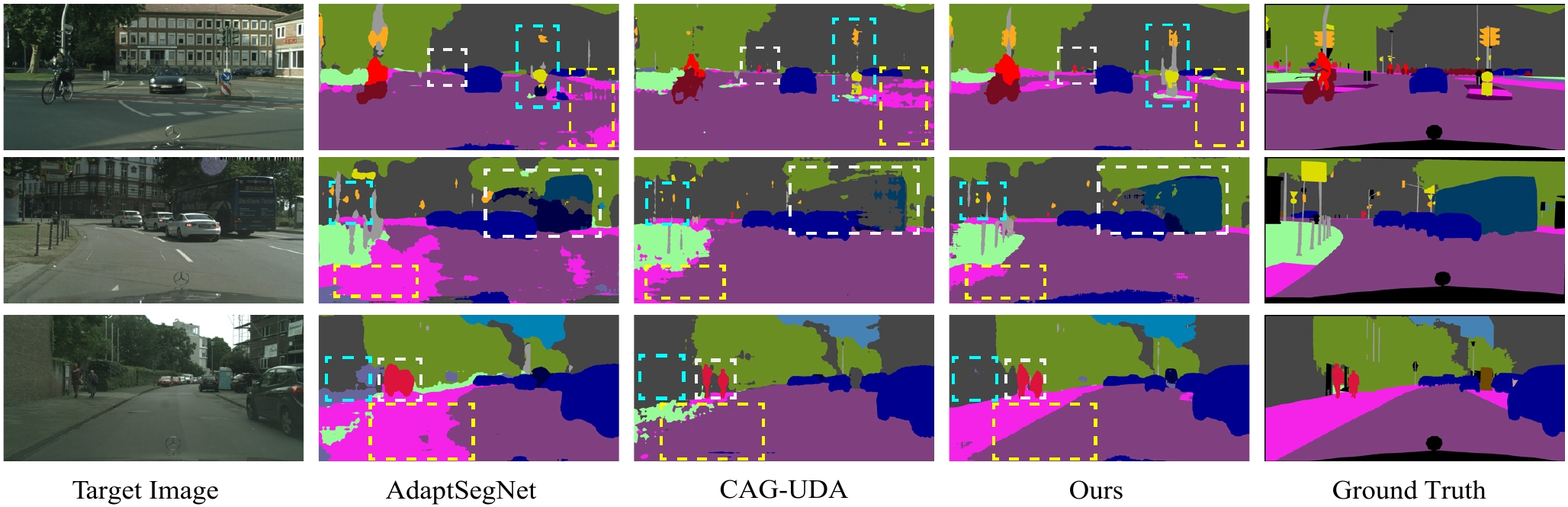}
    \caption{Qualitative results for GTA5 $\rightarrow$ Cityscapes task. For each target image, we show results with global alignment (AdaptSegNet), coarse-grained class alignment (CLAN) and our method, and the ground-truth label map.}
    \label{Fig_seg_map}
    \vspace{-4mm}
\end{figure*}

Specifically, for GTA5 $\rightarrow$ Cityscapes task, our method exceeds the global alignment based method, AdaptSegNet, by +9.8\% and +9.7\% for VGG-16 and ResNet-101 respectively. Our method also obtains +4.4\% improvements for Synscapes $\rightarrow$ Cityscapes task based on ResNet-101. Next, for SYNTHIA $\rightarrow$ Cityscapes task, our method compares favorably with the others. Compared with other class alignment approaches (CLAN, SSF-DAN, CAG-UDA, SIM, FADA etc.), a general gain of over 1.0\% is witnessed. Moreover, our method also performs comparable to or even better than the strong baselines, i.e., self-supervision~\cite{li2019bidirectional,lian2019pycda,pan2020unsupervised,vu2019advent,zou2019confidence} and pixel association~\cite{kang2020pixel,pixmatch2021_CVPR} approaches. These results reflect that the way of considering the per-pixel discriminative representation learning benefits the adaptation capability. Qualitative results for GTA5 $\rightarrow$ Cityscapes task are presented at Figure~\ref{Fig_seg_map}, verifying that our method also brings a significant visual improvement. More qualitative results are provided in Appendix~\ref{appendix:qualitative}.

\begin{figure*}
    \centering
    \includegraphics[width=\textwidth]{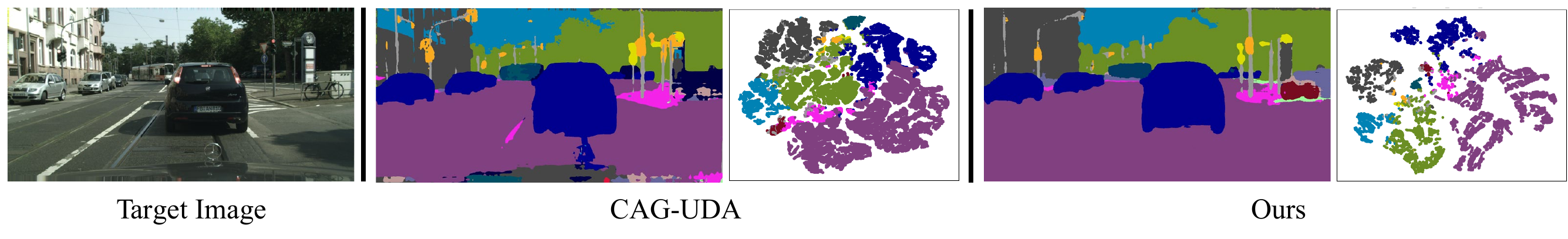}
    \caption{Comparisons of segmentation result and t-SNE visualization~\cite{maaten2008visualizing} between CAG-UDA~\cite{zhang2019category} and our method.}
    \label{Fig_tsne}
    \vspace{-3mm}
\end{figure*}

\begin{table*}[htbp]
    \begin{center}
        \caption{Experimental results for SYNTHIA $\to$ Cityscapes. mIoU$^{*}$ denotes the mean IoU of 13 classes, excluding the classes with $^{*}$.} \label{table:synthetic}
        \resizebox{\textwidth}{!}{
            \begin{tabular}{c|l c c c c c c c c c c c c c c c c|c|c}
                \toprule[1.2pt]
                Backbone & Method & \rotatebox{60}{road} & \rotatebox{60}{side.} & \rotatebox{60}{buil.} & \rotatebox{60}{wall$^{*}$} & \rotatebox{60}{fence$^{*}$} & \rotatebox{60}{pole$^{*}$} & \rotatebox{60}{light} & \rotatebox{60}{sign} & \rotatebox{60}{veg} & \rotatebox{60}{sky} & \rotatebox{60}{pers.} & \rotatebox{60}{rider} & \rotatebox{60}{car}&  \rotatebox{60}{bus} & \rotatebox{60}{mbike} & \rotatebox{60}{bike} & mIoU & mIoU$^{*}$ \\
                \hline
                \multirow{8}{*}{VGG-16}
                &AdvEnt~\cite{vu2019advent} & 67.9 & 29.4 & 71.9 & 6.3& 0.3& 19.9& 0.6& 2.6& 74.9& 74.9& 35.4& 9.6& 67.8& 21.4& 4.1& 15.5& 31.4& 36.6\\
                & CrCDA~\cite{huang2020contextual} & 74.5 & 30.5 & 78.6 & 6.6 & \bf 0.7 & 21.2 & 2.3 & 8.4 & 77.4 & 79.1 & 45.9 & 16.5 & 73.1  &  24.1 & 9.6 & 14.2 & 35.2 & 41.1 \\
                &CBST~\cite{zou2018unsupervised} & 69.6 & 28.7 & 69.5 & \bf 12.1 & 0.1 & 25.4 & \bf 11.9 & 13.6 & \bf 82.0 & 81.9 & 49.1 & 14.5 & 66.0 & 6.6 & 3.7 & \bf 32.4 & 35.4 & 36.1 \\
                &AdaptSegNet~\cite{tsai2018learning} & 78.9 & 29.2 & 75.5 & - & - & - & 0.1 & 4.8 & 72.6 & 76.7 & 43.4 & 8.8 & 71.1 & 16.0 & 3.6 & 8.4 &- & 37.6 \\
                &CLAN~\cite{luo2019taking} & 80.4 & 30.7 &74.7&-&-&-& 1.4& 8.0& 77.1& 79.0& 46.5& 8.9& 73.8& 18.2& 2.2& 9.9&-& 39.3\\
                &APODA~\cite{YangXLQSLL20} & 82.9 & 31.4 & 72.1 &-&-&-& 10.4 & 9.7 & 75.0 & 76.3 & 48.5 & 15.5 & 70.3 & 11.3 & 1.2 & 29.4 & - & 41.1 \\
                &FADA~\cite{wang2020class} & 80.4 & 35.9 & \bf 80.9 & 2.5 & 0.3 & \bf 30.4 & 7.9 & \bf 22.3 & 81.8 & \bf 83.6 & 48.9 & \bf 16.8 & 77.7 & 31.1 & 13.5 & 17.9 & 39.5 & 46.0 \\
                &\cellcolor{Gray}Ours & \cellcolor{Gray}\bf 86.6 & \cellcolor{Gray}\bf 41.0 & \cellcolor{Gray}79.5 & \cellcolor{Gray}1.7 & \cellcolor{Gray}0.2 & \cellcolor{Gray}28.5 & \cellcolor{Gray}0.0 & \cellcolor{Gray}6.8 & \cellcolor{Gray}80.5 & \cellcolor{Gray}82.7 & \cellcolor{Gray}\bf 54.8 & \cellcolor{Gray}14.5 & \cellcolor{Gray}\bf 83.3 & \cellcolor{Gray}\bf 36.3 & \cellcolor{Gray}\bf 19.7 & \cellcolor{Gray}32.3 & \cellcolor{Gray}\bf 40.5 & \cellcolor{Gray}\bf 48.3 \\         
                \hline
                \hline
                \multirow{22}{*}{ResNet-101}
                &Source Only & 56.8 & 21.5 & 75.5 & 5.3 & 0.1 & 26.2 & 10.3 & 13.8 & 77.2 & 73.2 & 53.6 & 15.6 & 77.1 & 30.3 & 10.9 & 17.5 & 35.3 & 41.0 \\
                &AdaptSegNet~\cite{tsai2018learning} & 79.2 & 37.2 & 78.8 & 10.5 & 0.3 & 25.1 & 9.9 & 10.5 & 78.2 & 80.5 & 53.5 & 19.6 & 67.0 & 29.5 & 21.6 & 31.3 & 39.5 & 45.9 \\ 
                &AdvEnt~\cite{vu2019advent} & 87.0 & 44.1 & 79.7 & 9.6 & 0.6 & 24.3 & 4.8 & 7.2 & 80.1 & 83.6 & 56.4 & 23.7 & 72.7 & 32.6 & 12.8 & 33.7 & 40.8 & 47.6 \\
                &CLAN~\cite{luo2019taking} & 81.3 & 37.0 & 80.1 & - &- & - & 16.1 &13.7 &78.2 & 81.5 & 53.4 & 21.2 & 73.0 & 32.9 & 22.6 & 30.7 & - & 47.8 \\
                &IntraDA~\cite{pan2020unsupervised} & 84.3 & 37.7 & 79.5 & 5.3 & 0.4 & 24.9 & 9.2 & 8.4 & 80.0 & 84.1 & 57.2 & 23.0 & 78.0 & 38.1 & 20.3 & 36.5 & 41.7 & 48.9 \\
                &CBST~\cite{zou2018unsupervised} & 68.0 & 29.9 & 76.3 & 10.8 & 1.4 & 33.9 & 22.8 & 29.5 & 77.6 & 78.3 & 60.6 & 28.3 & 81.6 & 23.5 & 18.8 & 39.8 & 42.6 & 48.9 \\
                &SSF-DAN~\cite{du2019ssf-dan} & 84.6 & 41.7 & 80.8 & - &- & - & 11.5 & 14.7 & 80.8 & 85.3 & 57.5 & 21.6 & 82.0 & 36.0 & 19.3 & 34.5 & - & 50.0 \\
                & CrCDA~\cite{huang2020contextual} & 86.2 & 44.9 & 79.5 & 8.3 & 0.7 & 27.8 & 9.4 & 11.8 & 78.6 & \bf 86.5 & 57.2 & 26.1 & 76.8 & 39.9 & 21.5 & 32.1 & 42.9 & 50.0 \\
                &CRST~\cite{zou2019confidence} & 67.7 & 32.2 & 73.9 & 10.7 & \textbf{1.6} & \bf 37.4 & 22.2 & \bf 31.2 & 80.8 & 80.5 & 60.8 & 29.1 & 82.8 & 25.0 & 19.4 & 45.3 & 43.8 & 50.1 \\                
                &BDL~\cite{li2019bidirectional} & 86.0 & 46.7 & 80.3 & - & - & - & 14.1 & 11.6 & 79.2 & 81.3 & 54.1 & 27.9 & 73.7 & 42.2 & 25.7 & 45.3 & - & 51.4 \\
                & WeakSegDA~\cite{Paul_WeakSegDA_ECCV20} & 92.0 & 53.5 & 80.9 & 11.4 & 0.4 & 21.8 & 3.8 & 6.0 & 81.6 & 84.4 & 60.8 & 24.4 & 80.5 & 39.0 & 26.0 & 41.7 & 44.3 & 51.9 \\
                &SIM~\cite{wang2020differential} & 83.0 & 44.0 & 80.3 & - &- & - & 17.1 & 15.8 & 80.5 & 81.8 & 59.9 & \textbf{33.1} & 70.2 & 37.3 & 28.5 & 45.8 & - & 52.1 \\
                &FDA~\cite{yang2020fda} &  79.3 & 35.0 & 73.2 & - & - &  - &  19.9 &  24.0 & 61.7 & 82.6 &  61.4 &  31.1 &  83.9 & 40.8 & \bf 38.4 & 51.1 & - &52.5 \\
                &FADA~\cite{wang2020class} & 84.5 & 40.1 & 83.1 & 4.8 & 0.0 & 34.3 & 20.1 & 27.2 & 84.8 & 84.0 & 53.5 & 22.6 & 85.4 & 43.7 & 26.8 & 27.8 & 45.2 & 52.5 \\ 
                &CAG-UDA~\cite{zhang2019category} & 84.8 & 41.7 & \bf 85.5 & - & - & - & 13.7 & 23.0 & \bf 86.5 & 78.1 & \bf 66.3 & 28.1 & 81.8 & 21.8 & 22.9 & 49.0 & - & 52.6 \\
                & CCM~\cite{li2020content} & 79.6 & 36.4 & 80.6 & 13.3 & 0.3 & 25.5 & 22.4 & 14.9 & 81.8 & 77.4 & 56.8 & 25.9 & 80.7 & 45.3 & 29.9 & \bf 52.0 & 45.2 & 52.9 \\
                &APODA~\cite{YangXLQSLL20} & 86.4 & 41.3 & 79.3 & - &- & - & 22.6 & 17.3 & 80.3 & 81.6 & 56.9 & 21.0 & 84.1 & 49.1 & 24.6 & 45.7 & - & 53.1 \\
                & LDR~\cite{yang2020label-driven} & 85.1 & 44.5 & 81.0 & - & - & -&16.4&15.2&80.1&84.8&59.4&31.9&73.2&41.0&32.6&44.7&- &53.1 \\
                &PLCA~\cite{kang2020pixel} & 82.6 & 29.0 & 81.0 & 11.2 & 0.2 & 33.6 & \bf 24.9 & 18.3 & 82.8 & 82.3 & 62.1 & 26.5 & 85.6 & 48.9 & 26.8 & 52.2 & 46.8 & 54.0 \\
                & PixMatch~\cite{pixmatch2021_CVPR} & \bf 92.5 & \bf 54.6 & 79.8 & 4.8 & 0.1 & 24.1 & 22.8 & 17.8 & 79.4 & 76.5 & 60.8 & 24.7 & 85.7 & 33.5 & 26.4 & \bf 54.4 & 46.1 & \bf 54.5 \\
                &\cellcolor{Gray}Ours & \cellcolor{Gray}86.9 & \cellcolor{Gray}43.2 & \cellcolor{Gray}81.6 & \cellcolor{Gray}\bf 16.2 & \cellcolor{Gray}0.2 & \cellcolor{Gray}31.4 & \cellcolor{Gray}12.7 & \cellcolor{Gray}12.1 & \cellcolor{Gray}83.1 & \cellcolor{Gray}78.8 & \cellcolor{Gray}63.2 & \cellcolor{Gray}23.7 & \cellcolor{Gray}\bf 86.9 & \cellcolor{Gray}\bf 56.1 & \cellcolor{Gray}33.8 & \cellcolor{Gray}45.7 & \cellcolor{Gray}\bf 47.2 & \cellcolor{Gray} 54.4 \\
                \bottomrule[1.2pt]
                \end{tabular}
        } 
        \vspace{-6mm}
    \end{center}
\end{table*}

\textbf{Results on tail classes.} We show the class distribution of Cityscapes validation set in Appendix~\ref{appendix:tail_class} and the tail classes are highlighted with blue in Table~\ref{table:gta} and Table~\ref{table:synscapes}. In essence, our method can achieve better performance at the majority of tail classes, demonstrating its superiority in mitigating the conditional domain shift. As mentioned earlier, the prior work CAG-UDA~\cite{zhang2019category} naively maximizes connections between pixels and the fixed category centroids whereas our method contrastively encourages the representation of each pixel in feature space to be close to the corresponding semantic prototype while being away from other prototypes. In Table~\ref{table:gta}, we can see that our method consistently performs better in terms of both mIoU and mIoU$_{tail}$. To further answer why our method are effective, we study from the t-SNE visualization~\cite{maaten2008visualizing} perspective. And we measure the high-dimensional pixel representations of prior work~\cite{zhang2019category} and ours to a 2D space with t-SNE shown in Figure~\ref{Fig_tsne}. The comparison results further prove that our method is more discriminative and consistent at a fine-grained level.

\begin{table*}[t]
    \centering
    \caption{Ablation of the proposed loss functions.}
    \label{table:ablation}
    \begin{tabular}{l|ccc|c|cc}
        \toprule[1.0pt]
        \multirow{2}{*}{Method} & \multirow{2}{*}{$\mathcal{L}_{cl}^s$} & \multirow{2}{*}{$\mathcal{L}_{cl}^t$} & \multirow{2}{*}{$\mathcal{L}_{ssl}$}  & GTA5 $\rightarrow$ Cityscapes & \multicolumn{2}{c}{SYNTHIA $\rightarrow$ Cityscapes} \\
          & & & & mIoU & mIoU & mIoU*  \\
         \hline
        \multirow{2}{*}{Baseline~\cite{tsai2018learning}} & {} & {} & {} & 41.4 & 39.5 & 45.9 \\
        & {} & {} & {\checkmark} & 44.5 & 42.3 & 49.1  \\ 
         \hline
         \multirow{3}{*}{Ours} 
         {} & {\checkmark} & {} & {} & 43.9 & 40.8 & 47.5 \\
         {} & {\checkmark} & {\checkmark} & {} & 49.2 & 45.8 & 52.6 \\
         {} & {\checkmark} & {\checkmark} & {\checkmark} & \bf 52.1 & \bf 47.2 & \bf 54.4 \\
        \bottomrule[1.0pt]
    \end{tabular} 
    \vspace{-4mm}
    
\end{table*}

\subsection{Ablation studies}
\label{sec:ablation}
We present comprehensive studies to evaluate the contribution of each component of our approach in Table~\ref{table:ablation}. First of all, we follow the work of~\cite{tsai2018learning} and employ global alignment in the output space, which serves as the baseline of our work. The results are shown in the first rows. 

{\bf Effect of contrastive loss.} Applying the proposed contrastive loss on the source domain ($\Lm_{cl}^s$) can outperform global alignment, which demonstrates enhancing intra-/inter-class affinities in the source domain is helpful for discriminative representation learning and also promotes the adaptability of segmentation model. To reduce the domain gap between domains, we further exploit the proposed contrastive loss on the target domain ($\Lm_{cl}^s + \Lm_{cl}^t$) and it contributes +{7.8}\% of mIoU and +{6.3}\% of mIoU for GTA5 $\rightarrow$ Cityscapes and SYNTHIA $\rightarrow$ Cityscapes respectively, which makes it comparable with other approaches. 

{\bf Effect of self supervised loss.} When only applying self supervision loss ($\mathcal{L}_{ssl}$) to the baseline, it leads to a slight gain. Naturally, we also employ $\mathcal{L}_{ssl}$ to explore the great veiled potentials of our method. This yields significant improvements, which benefits from explicitly enforcing intra-class pixel representations closer and inter-class pixel representations further apart.

\begin{table*}[t]
    \centering
    \begin{minipage}{0.48\textwidth}
        \centering
        \caption{Hyperparameter study on $\alpha$.}
        \label{table:parameter_alpha}
        \resizebox{\textwidth}{!}{
            \begin{tabular}{c|cccccc}
                \toprule[1.0pt]
                $\alpha$ & 0.0 & \cellcolor{Gray}0.1 & 0.2 & 0.5 & 0.8 & 1.0 \\
                \hline
                mIoU & 51.1 & \cellcolor{Gray}\bf52.1 & 52.1 & 52.0 & 51.7 & 48.6 \\
                \bottomrule[1.0pt]
            \end{tabular}
        }
    \end{minipage}
    \quad
    \begin{minipage}{0.48\textwidth}
        \centering
        \caption{Hyperparameter study $\lambda$.}
        \label{table:parameter_lambda}
        \resizebox{\textwidth}{!}{
            \begin{tabular}{c|cccccc}
                \toprule[1.0pt]
                $\lambda$ & 0.01 & 0.1 & 0.5 & \cellcolor{Gray}1.0 & 2.0 & 10.0 \\
                \hline
                mIoU & 50.3 & 51.4 & 51.9 & \cellcolor{Gray}\bf 52.1 & 51.8 & 50.1 \\
                \bottomrule[1.0pt]
            \end{tabular}
        }
    \end{minipage}
\end{table*}

{\bf Effect of hyperparameters.}
We conduct a study to tune the proper values of the hyperparameters $\alpha$ and $\lambda$ in our experiments. The selected hyperparameters for GTA5 $\rightarrow$ Cityscapes task are listed in Table~\ref{table:parameter_alpha} and Table~\ref{table:parameter_lambda}. Our method is able to achieve consistent performance within a wide range of $\alpha$ and $\lambda$. It worth noting that updated semantic prototypes gives better performance than fixed ones ($\alpha$ = 1.0).

\begin{figure*}[t]
    \centering
    \includegraphics[width=\textwidth]{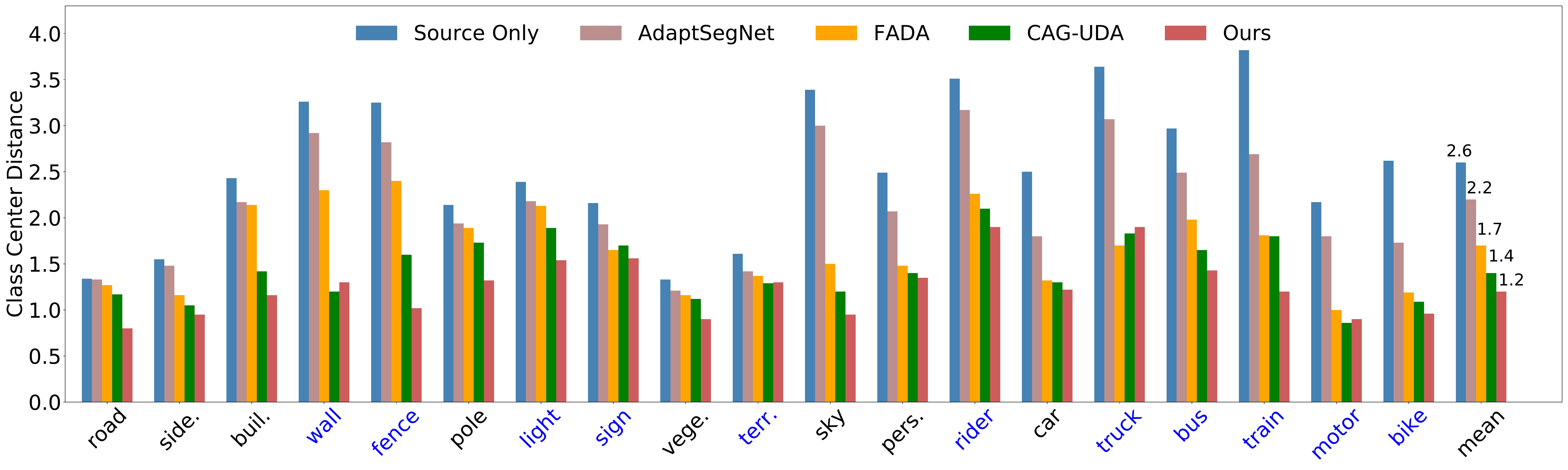}
    \caption{Quantitative analysis of the feature discrimination. For each class, we show the Class Center Distance (CDD) as defined in~\cite{wang2020class}, where a low CCD means the representations of the same class are densely clustered  while distances between different classes are relatively large. Our method shows a better aligned structure in class-level compared with other methods.}
    \label{Fig_center_distance}
    \vspace{-3mm}
\end{figure*}

\subsection{Feature discriminability}
To verify our method is capable of driving the intra-class pixel representations closer and the inter-class pixel representations further apart, we follow the Class Center Distance (CCD) metric in~\cite{wang2020class} to better compare the superiority of our method with other state-of-the-arts. Specifically, we randomly select 1,000 source images and 1,000 target images to calculate the CCD and report the comparison results in Figure~\ref{Fig_center_distance}. The results of our method are significantly lower than those trained using AdaptSegNet~\cite{tsai2018learning}, FADA~\cite{wang2020class}, and CAG-UDA~\cite{zhang2019category}, especially in the tail classes. It implies that our method is effective in driving pixel representations towards the corresponding semantic prototype to reduce the domain divergence and enhance feature discriminability.

\section{Conclusions}
In this paper, we present a novel semantic prototype-based contrastive learning to improve the adaptability in semantic segmentation. In particular, we design a fine-grained class alignment with an elegant contrastive loss to enforce the network to approximate the positive-concentrated and negative-separated properties of cross-domain pixel representations. 
The proposed method performs favorably against previous state-of-the-arts on multiple cross-domain segmentation scenarios. We believe that our exploration can inspire more efforts in this area.

%
%
\bibliographystyle{splncs04}
\bibliography{egbib}

\clearpage

\section*{Appendix}
\addcontentsline{toc}{section}{Appendices}
\renewcommand{\thesubsection}{\Alph{subsection}}

\subsection{Implementation details}
\label{appendix:implementation}

\begin{figure*}
   \centering
   \includegraphics[width=0.8\textwidth]{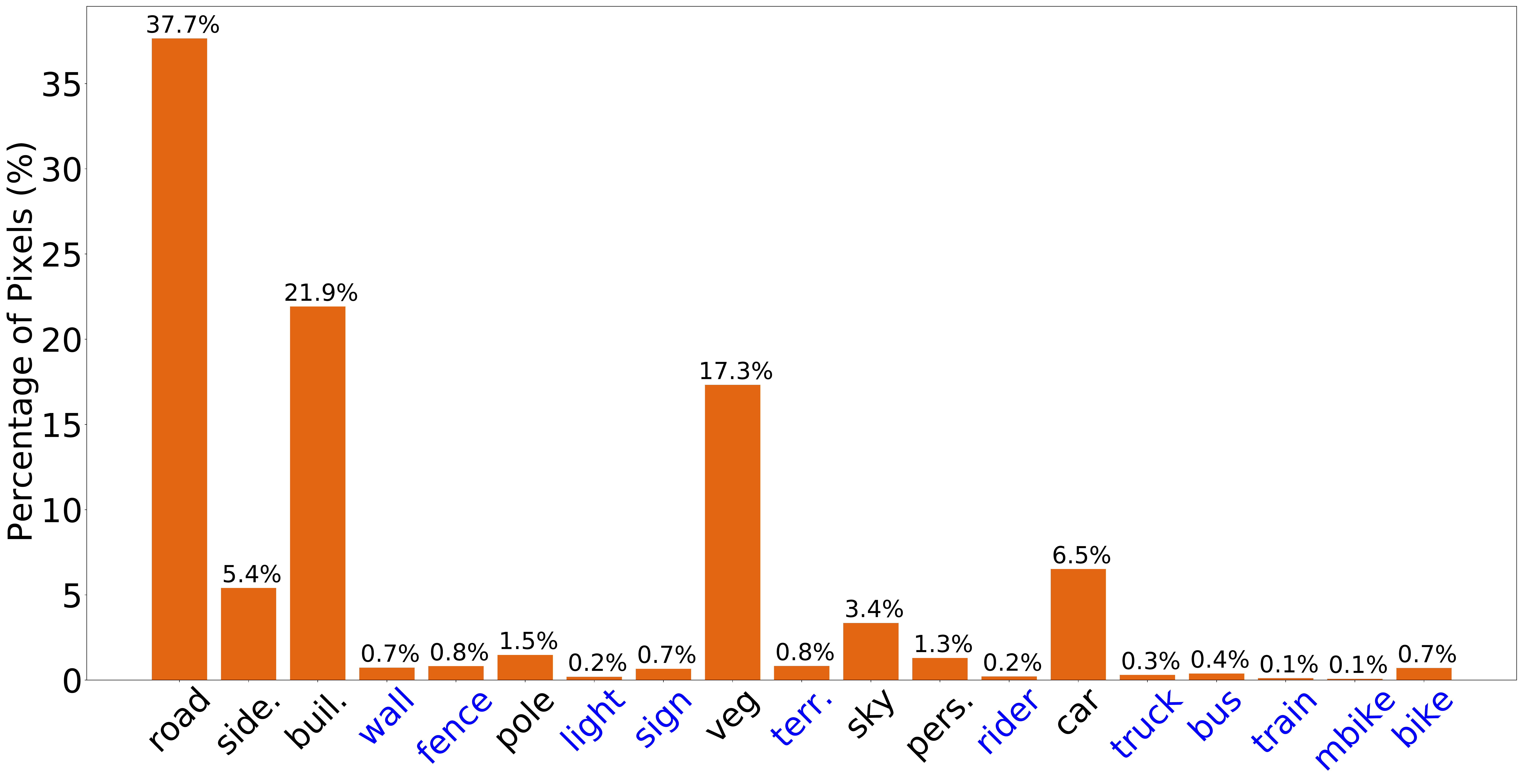}\vspace{-2mm}
   \caption{Class distribution on Cityscapes validation set. The tail classes are highlighted in \textcolor{blue}{blue}.}
   \label{Fig_tail_class}
\end{figure*}

\subsection{Class Distribution on Cityscapes Validation Set}
\label{appendix:tail_class}
Figure~\ref{Fig_tail_class} shows the class distribution of Cityscapes validation set. In this paper, those classes whose percentage of pixels is less than $1\%$ are defined as the tail classes. We can see that 11 out of 19 classes are the tail classes, and they are wall, fence, light, sign, terr., rider, truck, bus, train, mbike, bike.

\subsection{More Qualitative Comparisons}
\label{appendix:qualitative}
\begin{figure*}
   \centering
   \includegraphics[width=0.95\textwidth]{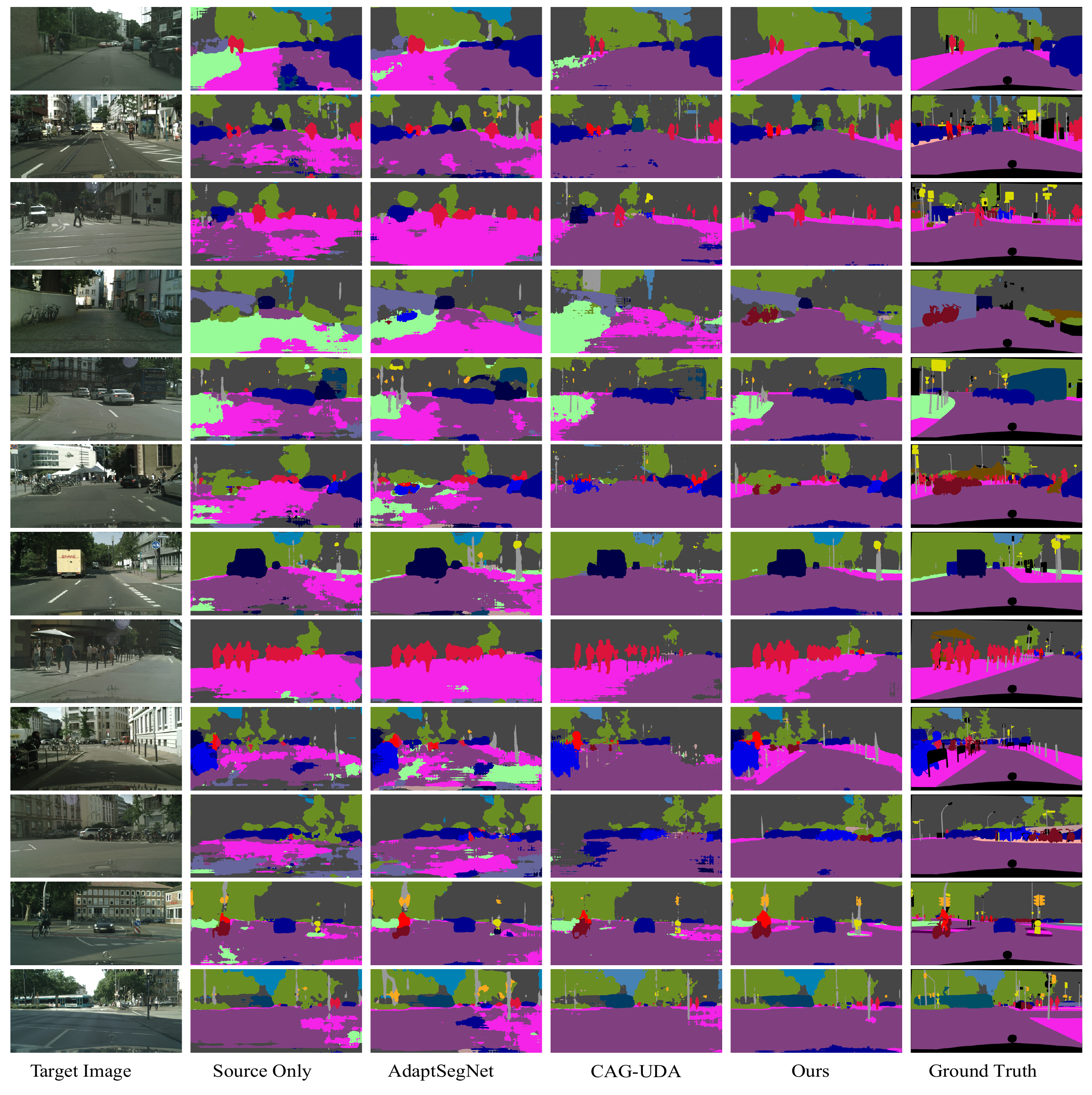}
   \vspace{-3mm}
   \caption{Example results of adapted segmentation for GTA5 $\rightarrow$ Cityscapes task. For each target image, we show segmentation results with Source Only, global alignment (AdaptSegNet~\cite{tsai2018learning}), coarse-grained class alignment (CAG-UDA~\cite{zhang2019category}), and our fine-grained class alignment and ground-truth label map.}
   \label{Fig_seg_map_gta}
\end{figure*}

\begin{figure*}
   \centering
   \includegraphics[width=\textwidth]{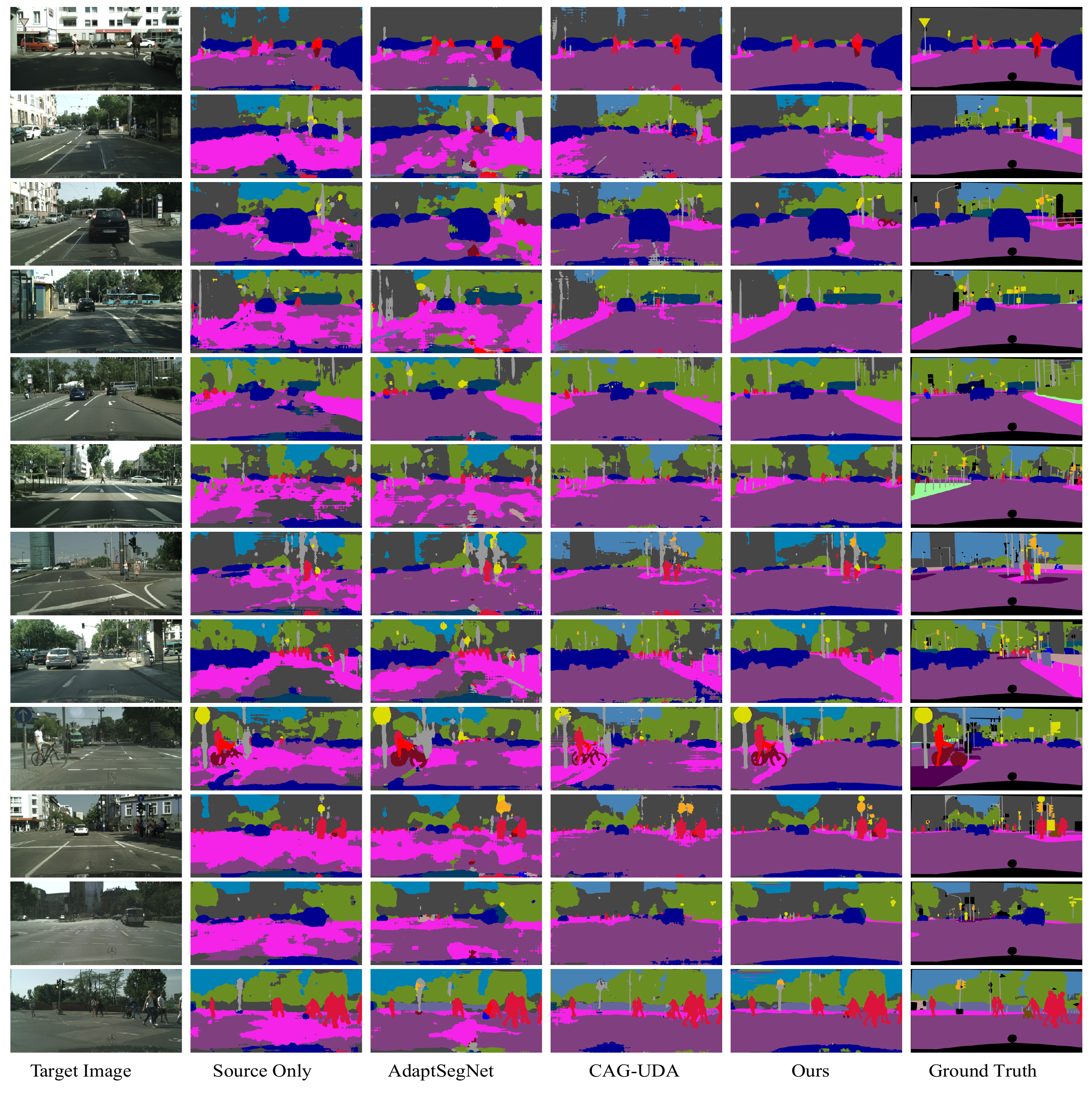}
   \caption{Example results of adapted segmentation for SYNTHIA $\rightarrow$ Cityscapes task. For each target image, we show results with Source Only, global alignment (AdaptSegNet~\cite{tsai2018learning}), coarse-grained class alignment (CAG-UDA~\cite{zhang2019category}), and our fine-grained class alignment and ground truth-label map.}
   \label{Fig_seg_map_synthia}
\end{figure*}

\begin{figure*}
   \centering
   \includegraphics[width=\textwidth]{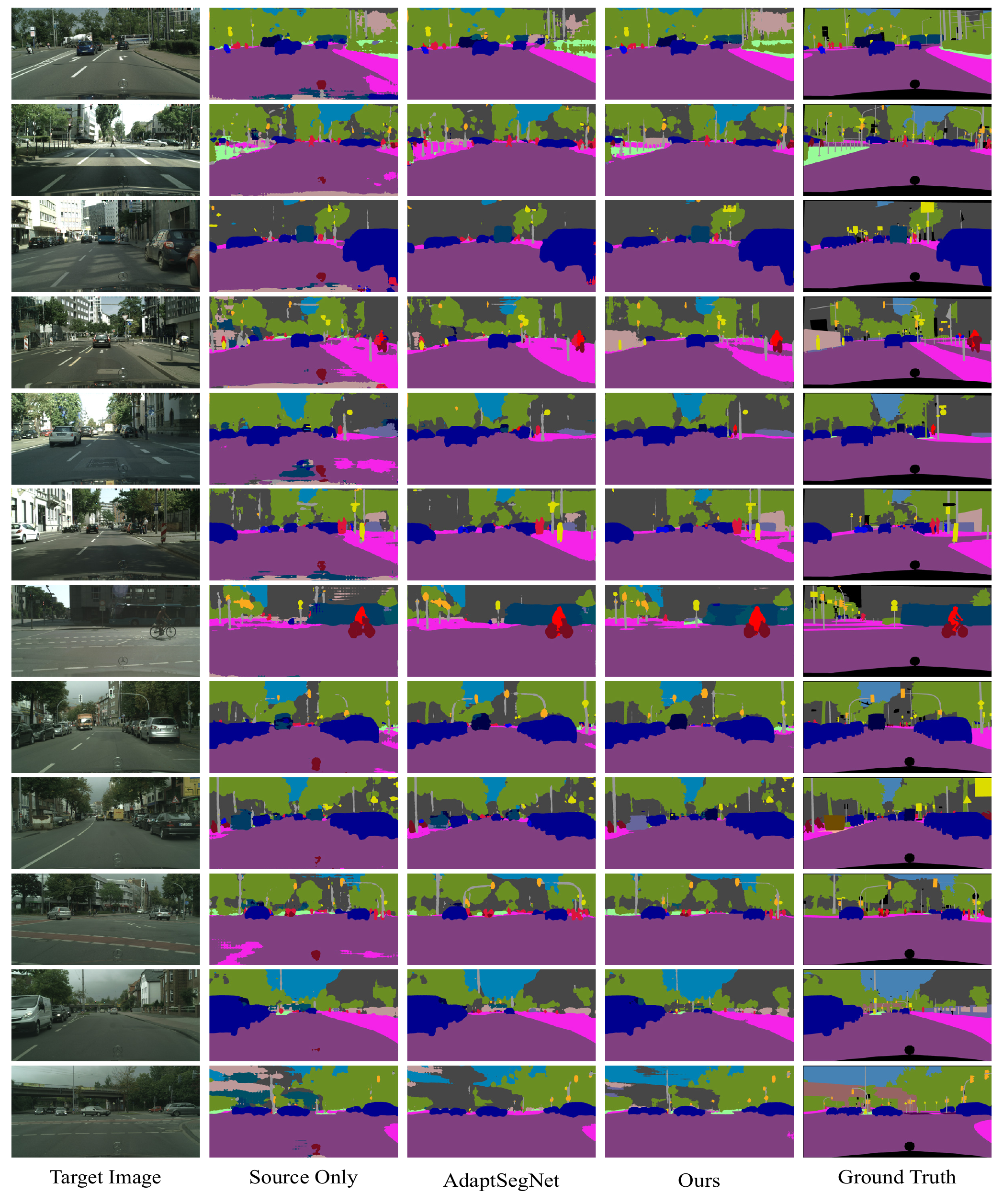}
   \caption{Example results of adapted segmentation for Synscapes $\rightarrow$ Cityscapes task. For each target image, we show results with Source Only, global alignment (AdaptSegNet~\cite{tsai2018learning}), and our fine-grained class alignment and ground-truth label map.}
   \label{Fig_seg_map_synscapes}
\end{figure*}

We present additional results for qualitative comparisons under various settings, including GTA5 $\rightarrow$ Cityscapes (Figure~\ref{Fig_seg_map_gta}), SYNTHIA $\rightarrow$ Cityscapes (Figure~\ref{Fig_seg_map_synthia}), and Synscapes $\rightarrow$ Cityscapes (Figure~\ref{Fig_seg_map_synscapes}). 
For GTA5 $\rightarrow$ Cityscapes and SYNTHIA $\rightarrow$ Cityscapes tasks, we visualize the segmentation results of the Source Only model, global alignment model (AdaptSegNet~\cite{tsai2018learning}), coarse-grained class alignment model (CAG-UDA~\cite{zhang2019category}) and our model. And for Synscapes $\rightarrow$ Cityscapes task, we show results of the Source Only model, global alignment model (AdaptSegNet~\cite{tsai2018learning}), and our model. The results predicted by our method are smoother and contain less spurious predictions than those predicted by other state-of-the-art approaches, demonstrating that with our fine-grained class alignment, it can bring a great visual improvement.

\subsection{Class Center Distance}
To further answer why our contrastive loss is effective, we study from the feature distribution perspective. We measure the Class Center Distance (CCD) metric~\cite{wang2020class} by taking intra-class and inter-class distances into account. The CCD for class $c$ is defined as follows:
\begin{small}
   \begin{align}
       CDD(c) = \frac{1}{C-1}\sum_{k=1,k\neq c}^{C}\frac{\frac{1}{|\Phi^c|}\sum_{x\in \Phi^c}{\left \| x -  \mu^c \right \| }^2}{{\left \| \mu^c - \mu^k \right \|}^2 }\,,
       \label{eq:cdd}
   \end{align}
\end{small}%
where $\mu^c$ is the semantic prototype of class $c$, $\Phi^c$ denotes the pixel set that contains all the pixel representations belonging to the $c$-th semantic class and $|\cdot|$ is the number of pixels in the set.

\end{document}